# Sequential Cross Mapping for Dynamical Causal Inference and Critical Early Warning

Jinling Yan[1], Shao-Wu Zhang[1*], Luonan Chen[2, 3,4*]

[1] MOE Key Laboratory of Information Fusion Technology, School of Automation, Northwestern Polytechnical University, Xi'an 710072, China
[2] School of Mathematical Sciences and School of AI, Shanghai Jiao Tong University, Shanghai 200240, China.
[3] Key Laboratory of Systems Health Science of Zhejiang Province, School of Life Science, Hangzhou Institute for Advanced Study, University of Chinese Academy of Sciences, Hangzhou 310024, China
[4] Tianfu Jincheng Laboratory, Chengdu 610212, China

* Corresponding author: Shao-wu Zhang; Luonan Chen
Email: zhangsw@nwpu.edu.cn; lnchen@sjtu.edu.cn

## Abstract

Detecting critical transitions in complex biological systems is essential for understanding cell-fate decisions and anticipating disease deterioration. However, conventional early-warning methods primarily rely on molecular fluctuations and statistical correlations, and indirect causal dependencies and nonlinear interactions further complicate the reconstruction of underlying causal networks. To address these challenges, We propose sequential cross mapping (SCM), a framework that integrates delay-coordinate reconstruction, sequential locally weighted linear mapping, and partial correlation to infer direct causal interactions while eliminating indirect associations. Based on the reconstructed causal networks, we develop CDE to quantify dynamical fluctuations of causal networks, thereby identifying potential early-warning signals and dynamic causal biomarkers. Applications to simulated dynamical systems and real biological datasets demonstrate that this method effectively reconstruct direct causal networks and detect critical states. Furthermore, CDE enables the identification of candidate dynamic causal biomarkers associated with key biological processes, including cell differentiation, proliferation, and PI3K–Akt signaling. Overall, our approach provides a promising strategy for characterizing the dynamic evolution of molecular causal networks and identifying critical transitions in complex biological systems.

## Introduction

Complex biological systems are governed by intricate interactions among genes, proteins, and signaling molecules, forming dynamic regulatory networks that maintain cellular functions and biological homeostasis [1-3]. During biological processes such as cellular differentiation, development, and disease progression, these systems may undergo critical transitions [4, 5], in which gradual changes in molecular regulation eventually leads to abrupt shifts in cellular states or phenotypes [6-8]. In particular, complex disease progression may involve transitions from relatively stable states to pathological states through a critical or pre-disease state. Detecting such critical states before pronounced deterioration occurs is therefore essential for understanding disease progression and developing effective early-warning strategies [9]. Importantly, critical transitions may be accompanied by substantial reorganization of molecular interactions, even when changes in individual molecular expression levels remain relatively subtle. Consequently, characterizing the dynamic evolution of regulatory networks may provide valuable information for identifying impending state transitions and understanding their underlying molecular mechanisms.

To detect critical states or anticipate sudden deterioration in complex biological systems, various early-warning approaches have been developed. Classical methods characterize declining dynamical stability through statistical indicators, such as increased variance and autocorrelation, particularly in systems exhibiting critical slowing down [4, 10]. Dynamical network biomarker (DNB) theory extends this perspective to molecular networks by identifying groups of molecules exhibiting increased expression fluctuations, stronger internal correlations, and weakened correlations with other molecules near a critical state [5]. DNB-based approaches have been successfully applied to the study of diabetes mellitus, hepatocellular carcinoma, breast cancer, and cell differentiation [11-14]. However, conventional DNB analysis generally requires multiple samples to estimate these statistical properties, limiting its application to individual-level disease prediction. To address this limitation, landscape DNBs (L-DNBs) [15], single-sample Kullback–Leibler divergence (sKLD) [16], hidden Markov models (HMMs) [17], and single-sample landscape entropy (SLE) [18] have been developed to identify critical states from individual samples with suitable reference information. More recently, network information gain (NIG) [19] and temporal network flow entropy (TNFE) [20] have further exploited network fluctuations and information-theoretic measures to characterize critical transitions. Complementary approaches based on potential landscapes quantify changes in biological state stability and energy barriers [21-23]. Alternatively, some machine learning-based approaches project data into latent spaces to further study the critical or bifurcation behaviors of dynamical systems. RC-TPD [24] identifies the tipping points in a CDS through delicately measuring the fluctuations emergent in the reservoir computing weights. FatePredictor [10] employs a dynamic unbalanced optimal transport method to reconstruct dynamic cell trajectories, based on which an ensemble deep learning model is utilized to predict the type of dynamics involved in a cell fate bifurcation during cellular processes. Although these approaches have demonstrated effectiveness in detecting critical states, many primarily rely on molecular fluctuations, statistical correlations, or system stability, without explicitly characterizing direct and directional interactions among variables.

Causal inference provides a means of characterizing directional dependencies among interacting variables in complex dynamical systems [9]. Various computational methods have been developed to infer causal interactions from observed time-series data, including Granger causality (GC) [25], transfer entropy (TE) [26], Neural Granger causality [27], convergent cross mapping (CCM) [28], cross map evaluation (CME) [29], cross map smoothness (CMS) [30], and partial cross mapping (PCM) [31]. Classical GC primarily characterizes linear predictive relationships, while TE and nonlinear extensions of GC broaden the analysis to nonlinear dependencies. Alternatively, CCM, CME, and CMS, rooted in delay-embedding theory and state-space reconstruction, provide approaches for analyzing causal interactions in nonlinear dynamical systems. However, causal dependencies inferred from observational data may contain both direct interactions and indirect influences transmitted through intermediate variables, potentially obscuring the underlying regulatory network. PCM addresses this problem by combining state-space reconstruction, mutual cross mapping, and partial correlation analysis to distinguish direct from indirect causal influences [31]. Embedding entropy and related methods further extend the inference of direct dynamical dependencies [32]. More recently, the IntDC framework extended dynamical causal inference from observational associations to interventional causal effects by deciphering interventional dynamical causality from non-interventional complex systems [33]. Building upon this framework, $IC^2$ further advances interventional dynamical causal inference by addressing latent confounders, enabling the identification of interventional causal relationships under partially observed dynamical systems [34]. These methods offer diverse and valuable perspectives for understanding causal interactions in complex dynamical systems. Inspired by these advances, integrating causal information into early-warning frameworks offers a more reliable and informative approach for detecting critical states and anticipating transitions in complex dynamical systems.

Recently, several studies have demonstrated the potential of integrating causal information into critical-transition detection. Sample-specific causality network entropy (SCNE) identifies pre-deterioration states by quantifying changes in individual-specific causal networks [35]. Causal network markers (CNMs) incorporate GC and TE to characterize directional interactions and construct early-warning indicators for critical transitions [36]. More recently, cell-specific causal network entropy (CCNE) has been developed to identify critical states by analyzing changes in molecular causal associations at the single-cell level [37]. These advances demonstrate that causal network information can complement conventional fluctuation- and correlation-based early-warning indicators. Nevertheless, accurately interpreting the evolution of causal networks remains challenging when inferred interactions are affected by indirect pathways and heterogeneous linear and nonlinear dynamics. In particular, changes in indirect dependencies may obscure the reorganization of direct causal interactions, making it difficult to identify the regulatory relationships most closely associated with critical transitions. However, indirect causal dependencies and heterogeneous nonlinear interactions may obscure changes in the underlying causal network. Therefore, reconstructing real causal networks and further quantifying their dynamical evolution may provide a more promising strategy for detecting critical transitions and identifying corresponding causal biomarkers.

To address these challenges, we propose SCM to infer direct causal associations from observed time series. Specifically, SCM combines delay-coordinate embedding, S-map, and partial correlation analysis to distinguish direct causal interactions from indirect dependencies mediated by other observed variables. Building upon the reconstructed causal networks, we further develop CDE to quantify causal information flow and its dynamical fluctuations in the network, thereby capturing early-warning signals and identifying dynamic causal biomarkers associated with critical transitions. We evaluate SCM using three-variable causal motifs, chaotic dynamical systems, an eight-node gene regulatory network, and simulated single-cell differentiation data. We further apply the proposed approach to embryonic hair-follicle development and mouse transcriptomic data of Alzheimer's disease to investigate its biological applicability. The results demonstrate that SCM effectively infers direct causal relationships while eliminating indirect associations, and CDE further identifies critical states and dynamic causal biomarkers associated with cell-fate transitions. In conclusion, this framework integrates dynamical causal inference with critical-transition analysis, providing a promising strategy for characterizing the dynamic evolution of molecular causal networks and identifying early-warning signals in complex biological systems.

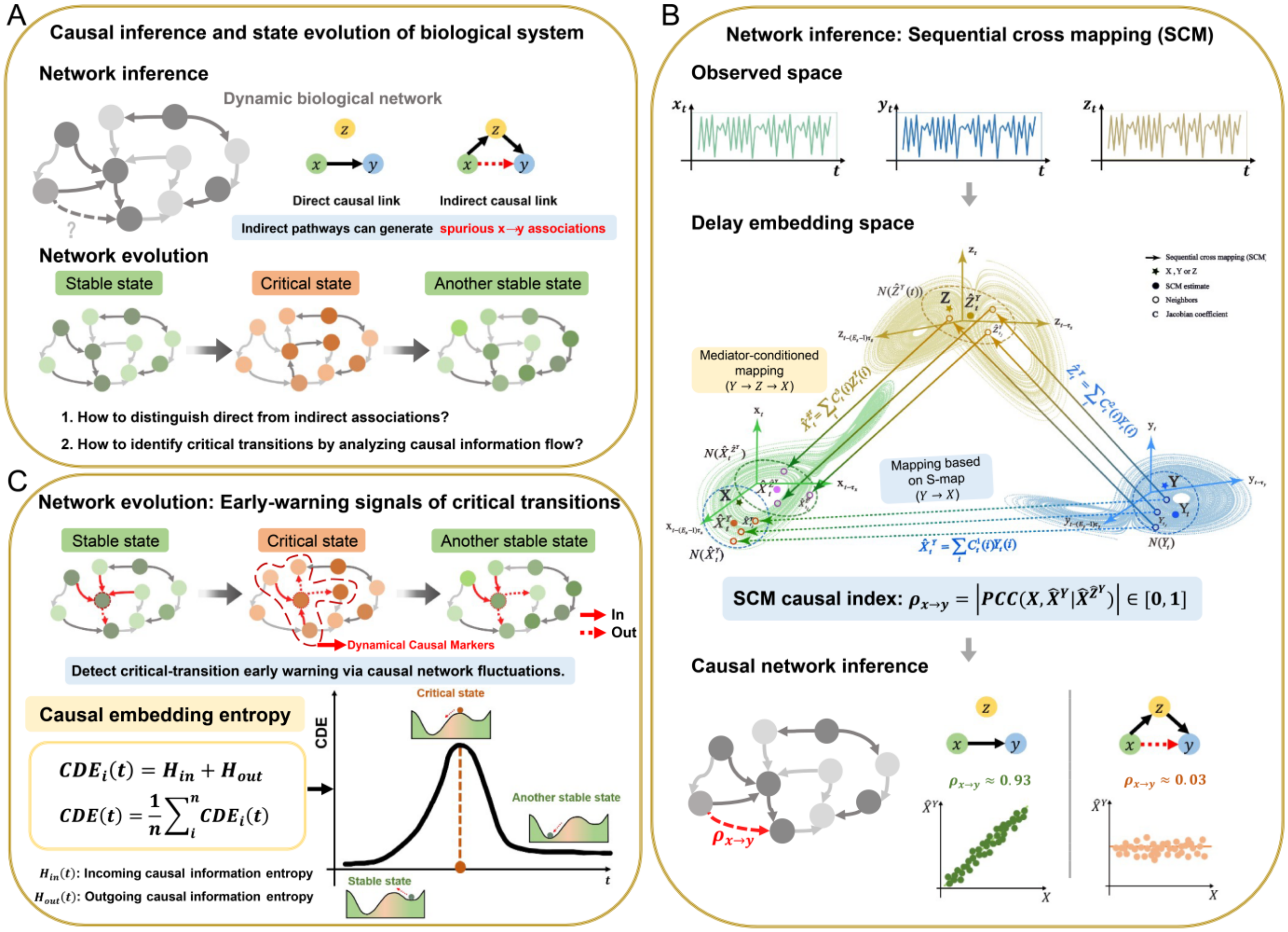

**Fig. 1. Framework of the Sequential cross mapping (SCM) method.** A. Problem formulation and biological background. Complex biological system consists of numerous interacting components, whose causal interactions collectively govern system dynamics. These interactions dynamically reorganize as the system evolves from stable to critical states, motivating the need for accurate causal network inference and detection of critical transitions. B. Biological network reconstruction based on causal inference. Direct causal inference using SCM. Multivariate time series are transformed into delay-coordinate embedding spaces, followed by S-map-based local state-space reconstruction. For a candidate causal relationship $x \to y$, SCM reconstructs $X$ directly from $Y$ and sequentially through the observed mediator $Z$. Partial correlation between the observed state $X$ and its direct cross-mapped estimate, conditioned on the mediator-based reconstruction, quantifies the remaining direct causal association while reducing indirect dependencies. C. Identify critical transitions in system states via the dynamical evoluton of network. CDE integrates incoming and outgoing causal information entropies to quantify the dynamical evolution of causal networks. The global CDE characterizes network-level fluctuations and provides potential early-warning signals of critical transitions, whereas node-level CDE profiles identify dynamic causal markers (DCMs) associated with transition-related regulatory changes.

# 2. Materials and methods

## 2.1. Problem formulation

A complex biological system consists of numerous interacting components, such as genes, proteins, and signaling molecules, whose causal interactions collectively govern system dynamics. However, reconstructing direct causal networks from observed time series remains challenging due to indirect dependencies mediated by other variables. Consider three interacting variables, $x$, $y$, and $z$. A direct causal interaction from $x$ to $y$, with $z$ evolving independently, can be represented as

$$\begin{cases} x_t = f(x_{t-1}) \\ y_t = f(y_{t-1}, x_{t-1}), \\ z_t = f(z_{t-1}) \end{cases} \tag{1}$$

In contrast, an indirect causal interaction from $x$ to $y$ through an intermediate variable $z$, denoted by $x \to z \to y$, can be expressed as

$$\begin{cases} x_t = f(x_{t-1}) \\ y_t = f(y_{t-1}, z_{t-1}). \\ z_t = f(z_{t-1}, x_{t-1}) \end{cases} \tag{2}$$

Both structures may give rise to observable associations between $x$ and $y$, even though their underlying causal mechanisms are fundamentally different, making it challenging to distinguish direct causal interactions from indirect influences transmitted through intermediate variables. Moreover, the dynamic evolution of causal networks is closely associated with changes in system states, making accurate identification of causal relationships essential for understanding system dynamics and detecting critical transitions. This brings up two fundamental challenges for biological systems (Fig.1A):

1) How to accurately distinguish direct from indirect causal interactions mediated by other observed variables?
2) How to identify critical transitions in system states by analyzing fluctuations in causal information flow?

## 2.2. Sequential cross mapping

To distinguish direct causal interactions from indirect dependencies, we develop Sequential Cross Mapping (SCM) by integrating delay-coordinate embedding, sequentially weighted global linear mapping (S-map) [38], and partial correlation analysis. The key idea is to quantify the agreement between an observed time series and its cross-mapped prediction through observed intermediate variables. Following this principle, SCM comprises three steps: cross-mapping reconstruction, two-stage cross mapping through a mediator, and quantification of direct causal relationships using partial correlation (Fig.1B).

Consider three variables, $x$, $y$, and $z$, forming the unidirectional causal chain $x \to z \to y$, as described in Eq. (2). Let $\boldsymbol{x} = \{x_t\}_{t=1}^{L}, \boldsymbol{y} = \{y_t\}_{t=1}^{L}, \boldsymbol{z} = \{z_t\}_{t=1}^{L}$ denote their observed time series of length $L$. According to Takens' delay-embedding theory, their corresponding shadow manifolds are reconstructed as

$$M_x = \{X_t\}_{t=r}^{L}, M_y = \{Y_t\}_{t=r}^{L}, M_z = \{Z_t\}_{t=r}^{L}.$$

where the delay-coordinate vectors are defined by

$$X_t = \left(x_t, x_{t-\tau}, x_{t-2\tau}, \dots, x_{t-(E-1)\tau}\right),$$
$$Y_t = \left(y_t, y_{t-\tau}, y_{t-2\tau}, \dots, y_{t-(E-1)\tau}\right),$$
$$Z_t = \left(z_t, z_{t-\tau}, z_{t-2\tau}, \dots, z_{t-(E-1)\tau}\right),$$

where $X_t, Y_t, Z_t \in \mathbb{R}^E$, $E$ is the embedding dimension, and $\tau$ is the time delay, and $r = 1 + (E-1)\tau$. In this study, $\tau$) is set to 1.

**(1) Cross-mapping reconstruction using S-map**

To assess the causal influence $x \to y$, SCM reconstructs the states of the cause variable $x$ from the shadow manifold $M_y$. For any pair of variables defined in Eq. (1) or Eq. (2), such as $x$ and $y$, let

$$\widehat{N}^Y(X_t) = \{X_{t_j} \mid Y_{t_j} \in N(Y_t)\},$$

where $N(Y_t)$ contains the $E+1$ nearest neighbor states of $Y_t$ on its corresponding shadow manifold $M_y$. Here, $E+1$ is the minimum number of points required to form a bounded simplex in an $E$-dimensional space. The time indices of neighborhood $Y_{t_j}$, denoted by $t_j$, are used to locate the corresponding cross-mapped neighbors $X_{t_j}$ on $M_X$, establishing a local mapping from $Y$ to $X$. Thus, the set $\widehat{N}^Y(X_t)$ represents the cross-mapped neighborhood derived from $N(Y_t)$, corresponding to the arrow from the manifold of $Y$ to that of $X$ in Fig. 1B. The dependence of $\widehat{N}^Y(X_t)$ on $N(Y_t)$ quantifies the causal influence of the dynamics of $x$ on those of $y$. This influence is measured by the correlation between $X_t$ and its cross-mapped estimate: $\hat{X}_t^Y = \mathbb{E}[\hat{N}^Y(X_t)]$. Here, $\hat{X}_t^Y$ is the estimate generated from $Y_t$, and $\mathbb{E}[\cdot]$ denotes an appropriately weighted average over all points in the specified set.

Unlike conventional cross mapping [28, 31] based on neighborhood-weighted averages, SCM employs S-map to estimate the cross-mapped states through locally weighted linear regression. The resulting cross-mapped estimate is given by

$$\hat{X}_t^Y = \sum_{i=0}^{E} C_t(i)\, Y_t(i). \tag{3}$$

Here, $C_t$ is the local mapping coefficient vector estimated by weighted linear regression. Excluding the intercept, its coefficients characterize a local linear approximation to the Jacobian of the $Y \to X$ reconstruction map in the embedding space. At each time $t$, $C_t$ is obtained by solving the following weighted least-squares problem using singular value decomposition (SVD):

$$AC_t = B. \tag{4}$$

The elements of $A \in \mathbb{R}^{n\times(E+1)}$ and $B \in \mathbb{R}^{n\times 1}$ are defined as $A_{ji} = \omega_{t_j} X_t(i), B_j = \omega_{t_j}\tilde{X}_{t_j}$, where $i = 0, \ldots, E, j = 1, \ldots, n$. Here, $\tilde{X}_t = X_{t+T_p}$ is the prediction target, $T_p$ is the prediction horizon, and $Y_{t_j}(0) = 1$ provides the bias term, corresponding to the intercept. The weight $\omega_{t_j}$ assigned to the $j$ −th neighboring point, $\omega_{t_j}$, depends on the Euclidean distance between the neighboring state $Y_{t_j}$ and the target state $Y_t$,

$$\omega_{t_j} = \exp\left\{-\frac{\theta \cdot \| Y_{t_j} - Y_t \|}{\sum_{t_j=1}^{E+1} \| Y_{t_j} - Y_t \|}\right\}, \theta \geq 0. \tag{5}$$

Here, $\theta$ controls the degree of localization. When $\theta = 0$, all neighboring vectors receive equal weights, yielding a global linear mapping. When $\theta > 0$, neighboring states closer to the predicted point $Y_t$ in state space receive larger weights. As $\theta$ increases, the model progressively shifts toward a local nonlinear mapping, thus to capture state-dependent nonlinear relationships. By repeating this procedure over the observed time series, SCM obtains the cross-mapped prediction $\hat{X}^Y$, which represents the information about $X$ reconstructed from the dynamics of $Y$. The agreement between $X$ and $\hat{X}^Y$ characterizes their overall cross-mapping dependence, which may contain both direct and indirect causal contributions.

**(2) Two-stage sequential cross mapping through a mediating variable**

To account for indirect causal dependencies arising from cascade structures in biological networks, SCM constructs a two-stage sequential cross-mapping procedure through an intermediate variable $z$.

First, the states of $Z$ are reconstructed from the shadow manifold $M_Z$ using the S-map procedure described above, yielding

$$\hat{Z}_t^Y = \sum_{i=0}^{E} C_t'(i)\, Y_t(i), \tag{6}$$

where $C_t' \in \mathbb{R}^{E+1}$ is the coefficient vector of the $Y \to Z$ mapping. As the sample size $L$ increases and the parameter $\theta$ is optimized, $\hat{Z}_t^Y$ asymptotically converges to the true dynamical state of $Z$, indicating that $Z$ can be effectively reconstructed from the historical information contained in $Y$.

Next, the same procedure is repeated using $X$ and the sequentially cross-mapped series $\hat{Z}^Y$ to obtain the mapping from $\hat{Z}^Y$ to $X$, denoted by $\hat{X}^{(\hat{Z}^Y)}$:

$$\hat{X}_t^{(\hat{Z}^Y)} = \sum_{i=0}^{E} C_t''(i)\, \hat{Z}_t^Y(i), \tag{7}$$

where $C_t'' \in \mathbb{R}^{E+1}$ is the coefficient vector of the $\hat{Z}^Y \to X$ mapping. The estimate $\hat{X}^{(\hat{Z}^Y)}$ characterizes the information about $X$ that can be reconstructed through the intermediate variable $Z$, corresponding to the indirect pathway $x \to z \to y$. Thus, $\hat{X}^Y$, obtained through the cross mapping from $Y$ to $X$ ($M_Y \to M_X$), captures the overall cross-mapping dependence between $x$ and $y$, whereas $\hat{X}^{(\hat{Z}^Y)}$, obtained through the two-stage mapping $M_Y \to M_Z \to M_X$, characterizes the mediator-associated component of this dependence. These two estimates are subsequently used in partial correlation analysis to quantify the remaining directional dependence after accounting for information reconstructed through mediator $z$.

**(3) Quantification of direct causal strength using partial correlation analysis**

Based on the direct and two-stage cross-mapped estimates, SCM introduces a partial-correlation-based causal index to quantify the directional dynamical dependence from $x$ to $y$ after conditioning on information reconstructed through $z$:

$$\rho_{x\to y} = \left|\mathrm{PCC}\left(X, \hat{X}^{Y} \,\middle|\, \hat{X}^{(\hat{Z}^{Y})}\right)\right|. \tag{8}$$

Here, $PCC(\cdot,\cdot|\cdot)$ denotes the partial correlation coefficient, which measures the correlation between the first two time series after removing their linearly explainable components associated with the conditioning variable. Accordingly, $\rho_{x\to y}$ quantifies the cross-mapping dependence between $X$ and $\hat{X}^{Y}$ after accounting for the indirect information reconstructed through $z$, providing an estimate of their direct causal association from $x$ to $y$. Notely, in this method, adjusting $\theta$ in S-map enables a continuous transition from a global linear description to a local nonlinear representation, allowing SCM to accommodate systems with different dynamical characteristics. Finally, SCM approximates the dynamical properties of the system by progressively estimating the local mapping matrix $C$ as the system evolves along its attractor trajectory. This enables further analysis of how causal strengths evolve with the system state, providing a potential basis for identifying key state transitions in dynamical systems. In complex biological networks, indirect dependencies may arise through multiple intermediate variables or causal pathways. Therefore, the above formulation can be extended to a system containing $m$ candidate mediators, $\boldsymbol{z} = \{z_1, z_2, \dots, z_m\}$. For each mediator $z_k$, the two-stage procedure yields a reconstructed series $\hat{X}^{(\hat{Z}_k^{Y})}$. Their combined contributions are accounted for through a multivariate partial correlation:

$$\rho_{x\to y}^{(\boldsymbol{z})} = \left|\mathrm{PCC}\left(X, \hat{X}^{Y} \,\middle|\, \hat{X}^{(\hat{Z}_k^{Y})}, \dots, \hat{X}^{(\hat{Z}_m^{Y})}\right)\right|. \tag{9}$$

This formulation jointly conditions on the reconstructed information associated with multiple observed mediators, reducing the risk of attributing indirect dependencies to direct interactions. For longer cascades involving successive intermediate variables, higher-order cross mappings can be constructed analogously and incorporated into the conditioning set.

## 2.3 Critical statistical properties of the causal index

To characterize the fluctuations of causal interactions as a dynamical system approaches a critical transition, we investigate the statistical properties of the SCM variables, $z_i$, $\hat{z}_i^{z_j}$, and $\hat{z}_i^{\hat{z}_r^{z_j}}$.

### 2.3.1 Critical statistical properties of the variables

Consider a discrete-time nonlinear dynamical system:

$$Z(t+1) = f(Z(t); P), \tag{10}$$

where $Z(t) = \left(z_1(t), z_2(t), \dots, z_n(t)\right)^{\mathrm{T}} \in \mathbb{R}^n$ is the state vector of $n$ gene expression levels at time $t$, $P \in \mathbb{R}^s$ denotes slowly varying driving factors (e.g., genetic or epigenetic factors), and $f$ is a general nonlinear function describing the system dynamics. Assume that the system has a stable fixed point $\bar{Z} = f(\bar{Z}; P)$, as $P$ approaches a critical value $P_c$, the eigenvalue of the Jacobian with the largest modulus approaches the unit circle. At $P = P_c$, the system undergoes a local change in stability at $\bar{Z}$, such as a codimension-one bifurcation, with the corresponding eigenvalue satisfying $|\lambda| = 1$. Near a stable fixed point $\bar{Z}$, local linearization and diagonalization give

$$Y(t+1) = \Lambda Y(t) + \zeta(t),$$

where $\boldsymbol{Y}(t) = S^{-1}(\boldsymbol{Z}(t) - \bar{\boldsymbol{Z}})$, $\Lambda = S^{-1}JS = \mathrm{diag}(\lambda_1, \dots, \lambda_n)$, and $\zeta(t)$is zero-mean Gaussian noise with covariance $\sigma_{ij} = \mathrm{Cov}(\zeta_i, \zeta_j)$. For simplicity, we consider a single real dominant eigenvalue $\lambda_1 \to 1$, while the remaining eigenvalues remain strictly within the unit circle. The stationary variance and covariance of the transformed variables satisfy

$$\mathrm{Var}(y_i) = \frac{\sigma_{ii}}{1-\lambda_i^2}, \qquad \mathrm{Cov}(y_i, y_j) = \frac{\sigma_{ij}}{1-\lambda_i\lambda_j}.$$

To examine the corresponding changes in the observed variables, consider three representative variables $z_i$, $z_j$, and $z_r$ corresponding to the cause, effect, and intermediate variables in SCM, respectively. and assume that $\Lambda$ is fully diagonalized. Using $Z(k) - \bar{Z} = SY(k)$, the observed variables can be written as

$$\begin{cases} z_i(k) = s_{i1}y_1(k) + \cdots + s_{in}y_n(k) + \bar{z}_i \\ z_j(k) = s_{j1}y_1(k) + \cdots + s_{jn}y_n(k) + \bar{z}_j \\ z_r(k) = s_{r1}y_1(k) + \cdots + s_{rn}y_n(k) + \bar{z}_r \end{cases}, \tag{11}$$

Accordingly, the variance of $z_i$ and its covariance with $z_j$ can be expressed as Eq. (S1.11) and Eq. (S1.12). Similar expressions can be obtained for $z_j$ and $z_r$. Since $y_1$ is associated with the dominant eigenvalue $\lambda_1$, if a variable $z_i$ is associated with $y_1$(i.e., $s_{i1} \neq 0$), its variance $\mathrm{Var}(z_i)$increases significantly as $\lambda_1 \to 1$. If both $z_i$ and $z_j$are associated with $y_1$ (i.e., $s_{i1} \neq 0$and $s_{j1} \neq 0$), the magnitude of $Cov(z_i, z_j)$ increases significantly as well. Conversely, if at least one of $z_i$and $z_j$ is independent of $y_1$, their covariance tends to a constant as $\lambda_1 \to 1$. These results indicate that variables associated with the dominant eigenmode exhibit amplified statistical fluctuations near criticality, providing the dynamical basis for investigating the critical properties of SCM-derived causal interactions.

### 2.3.2 Critical statistical properties of the causal index

To further characterize the fluctuations of causal strength near the critical transition (as $P \to P_c$), we analyze the statistical properties of the SCM causal index. For a candidate causal relationship $z_i \to z_j$, the causal index is determined by the original variable $z_i$, its cross-mapped estimate $\hat{z}_i^{z_j}$, and the two-stage cross-mapped estimate $\hat{z}_i^{\hat{z}_r^{z_j}}$,

$$\hat{z}_i^{z_j} = \sum_{l=0}^{E} C_t^{(i\leftarrow j)}(l) z_j^{(l)}, \hat{z}_i^{\hat{z}_r^{z_j}} = \sum_{l=0}^{E} C_t^{(i\leftarrow \hat{r}\leftarrow j)}(l) z_j^{(l)},$$

where $z_j^{(l)}$ denotes the $l$-th delay coordinate. Here, $C_t^{(i\leftarrow j)} \in \mathbb{R}^{E+1}$ and $C_t^{(i\leftarrow \hat{r}\leftarrow j)} \in \mathbb{R}^{E+1}$ are the local mapping coefficients from the delay-embedded state of $z_j$ to $z_i$ and the two-stage mapping $z_j \to \hat{z}_r^{(z_j)} \to z_i$, respectively. Based on these expressions, the variances and covariances among $z_i$, $\hat{z}_i^{z_j}$, and $\hat{z}_i^{\hat{z}_r^{z_j}}$ can be derived from the statistical properties of the original variables and their mapping coefficients (see Supplementary Information, Eqs. S2.13–S2.20). These quantities determine the pairwise Pearson correlations and the SCM causal index

$$PCC(z_i, \hat{z}_i^{z_j} | \hat{z}_i^{\hat{z}_r^{z_j}}) = \frac{PCC(z_i, \hat{z}_i^{z_j}) - PCC\left(z_i, \hat{z}_i^{\hat{z}_r^{z_j}}\right) \times PCC\left(\hat{z}_i^{z_j}, \hat{z}_i^{\hat{z}_r^{z_j}}\right)}{\sqrt{\left(1 - PCC\left(z_i, \hat{z}_i^{\hat{z}_r^{z_j}}\right)^2\right)\left(1 - PCC\left(\hat{z}_i^{z_j}, \hat{z}_i^{\hat{z}_r^{z_j}}\right)^2\right)}} \tag{12}$$

Therefore, the critical behavior of $PCC(z_i, \hat{z}_i^{z_j} | \hat{z}_i^{\hat{z}_r^{z_j}})$ depends on the joint statistical properties of the observed variable and its two cross-mapped estimates. Based on these statistical properties, the dynamical characteristics of $z_i$, $\hat{z}_i^{z_j}$, and $\hat{z}_i^{\hat{z}_r^{z_j}}$ near the critical point fall into eight cases, depending on whether $s_{i1}$, $s_{j1}$, and $s_{r1}$ are zero or non-zero (Supplementary Table 2-1). These cases condense into three asymptotic patterns for a causal link $z_i \to z_j$as $\lambda_1 \to 1^-$:

1) If both $z_i$ and $z_j$ are associated with $y_1$, the causal strength between them increases markedly and approaches 1. Specifically, as $\lambda_1 \to 1$, $PCC(z_i, \hat{z}_i^{z_j} | \hat{z}_i^{\hat{z}_r^{z_j}}) \to 1$.
2) If only one of $z_i$ and $z_j$ is associated with $y_1$, then as $\lambda_1 \to 1$, $PCC(z_i, \hat{z}_i^{z_j} | \hat{z}_i^{\hat{z}_r^{z_j}}) \to 0$.
3) If neither $z_i$ nor $z_j$ is associated with $y_1$, then throughout the dynamical evolution of the system, regardless of whether a critical transition occurs, $PCC(z_i, \hat{z}_i^{z_j} | \hat{z}_i^{\hat{z}_r^{z_j}}) \to C$, where $C$ denotes a constant.

Therefore, all variables $z_i$ with $s_{i1} \neq 0$ form the **critical dominant variable group**, which will first perform a significant signal when system experiences state transition, e.g., tipping point or bifurcation. The critical dominant variables capture key dynamical features of the underlying system: near the critical state, causal effects within the dominant group are significantly enhanced, causal relationships among non-dominant variables remain at a constant level, and causal effects between dominant and non-dominant variables are significantly reduced. Thus, the SCM causal index and its statistical properties provide the theoretical motivation for quantifying fluctuations in causal information flow and identifying dynamical causal markers associated with impending state transitions.

### 2.3.3 Causal embedding entropy (CDE)

Based on the critical statistical properties of the SCM causal index, we further develop Causal Embedding Entropy (CDE) to characterize information flows in causal networks and identify early-warning signals of critical transitions (Fig.1C). We first construct the causal network of system (2.9) using SCM, where nodes represent system variables and edges represent inferred direct causal associations. To quantify changes in the incoming and outgoing causal information associated with each variable, we define a composite indicator, CDE, as

$$CDE_i(t) = H_{in} + H_{out}, \tag{13}$$

where $H_{\text{in}}(t)$ and $H_{\text{out}}(t)$ are the incoming and outgoing causal information entropies of variable $z_i$ at time $t$, respectively, calculated as

$$H_{in}(t) = -\sum_{j}^{N_{in(i)}} z_i^{t_j}(t)\hat{\rho}_{j\to i}(t) \log\left(z_i^{t_j}(t)\hat{\rho}_{j\to i}(t)\right), \tag{14}$$

$$H_{out}(t) = -\sum_{k}^{N_{out(i)}} z_i(t)\hat{\rho}_{i\to k}(t) \log\left(z_i(t)\hat{\rho}_{i\to k}(t)\right), \tag{15}$$

The quantities $z_i(t)$ and $z_i^{t_j}(t)$ are the observed values of node $z_i$ and its $j$th incoming neighbor $z_i^{t_j}$, respectively, with $z_i^{t_j} \to z_i$. The weight $\hat{\rho}_{j\to i}(t)$ denotes the causal strength from $z_i^{t_j}$ to $z_i$ at time $t$, and $\hat{\rho}_{i\to k}(t)$ denotes the causal strength from $z_i$ to its outgoing neighbor $z_i^{t_k}$. The incoming-neighbor set is denoted by $N_{\text{in}}(i)$, with $z_i^{t_j} \in N_{\text{in}}(i)$; the outgoing-neighbor set is denoted by $N_{\text{out}}(i)$, with $z_i^{t_k} \in N_{\text{out}}(i)$ and $z_i \to z_i^{t_k}$.

The node-level indicator $\text{CDE}_i(t)$ quantifies fluctuations in the causal information associated with $z_i$ as the system approaches the critical point. It therefore provides an early-warning signal at the level of an individual variable. As the system approaches the critical state, $\text{CDE}_i(t)$ may exhibit one of the following three asymptotic behaviors:

$$\text{CDE}_i = \begin{cases} increase\ greatly\ (\infty) & if\ z_i, z_{t_j} \in DCMs \\ keep\ a\ constant\ (C) & if\ z_i, z_{t_j} \notin DCMs \\ decrease\ greatly\ (0) & if\ z_i\ or\ z_{t_j} \in DCMs \end{cases}, \tag{16}$$

As the system approaches criticality, these variables are driven by the dominant dynamics to exhibit pronounced dynamical fluctuations before other variables, and are thus defined as candidate Dynamical Causal Markers (DCMs). To quantify the early-warning signal at the system level, we average the node-level indicators over the network (Fig.1C):

$$CDE(t) = \frac{1}{n}\sum_{i}^{n} CDE_i(t), \tag{17}$$

where $n$ denotes the total number of variables in the system. As the system approaches a critical transition, the coordinated reorganization of directed causal interactions may produce a pronounced increase or local peak in the global CDE. Such changes provide potential early-warning signals, while the corresponding node-level CDE scores help identify the variables most strongly associated with critical state transitions. Building on this connection, the proposed method extends critical-state analysis from correlation-based networks to directed causal interaction networks By integrating SCM-derived causal strengths with causal embedding entropy, the DCB framework further characterizes the direction, strength, and dynamic reorganization of inferred causal interactions, providing complementary insights into the underlying network dynamics and early-warning signals of critical transitions.

### 2.4. Evaluation systems and interpretation of biological data

To systematically evaluate the proposed Sequential Cross Mapping (SCM) framework for direct causal inference and critical early warning, we conducted experiments on simulated dynamical systems and biological datasets. First, five three-variable causal motifs, including direct interactions, causal chains, feedback loops, common drivers, and colliders, were used to assess the ability of SCM to distinguish direct from indirect causal associations. The Lorenz and Rössler systems were then employed to investigate CDE-based early-warning signals across different dynamical regimes. An eight-node logistic regulatory network was then used to evaluate causal network reconstruction and node-level CDE fluctuations near criticality. To further assess its applicability to biological systems, we analyzed a SERGIO-simulated single-cell dataset comprising 700 cells, 100 genes, and seven cell types to explore causal network dynamics during cell differentiation [39]. We subsequently applied SCM to two real biological datasets, mouse embryonic hair-follicle development [40] and the AD11 mouse transcriptomic dataset [41], to reconstruct causal network and identify potential dynamical causal markers during development and disease progression. Finally, Gene Ontology (GO) and Kyoto Encyclopedia of Genes and Genomes (KEGG) enrichment analyses were performed to investigate the biological functions of the identified markers.

## 3. Results

### 3.1. Direct causal inference and early warning in simulated dynamical systems

To evaluate the ability of SCM to reconstruct causal networks and identify critical points, we applied it to the following three-species dynamical system,

$$\begin{cases} x_t = x_{t-1}\left[\alpha_x - \alpha_x x_{t-1} - \beta_{yx} y_{t-1} - \beta_{zx} z_{t-1}\right] + \epsilon_{x,t} \\ y_t = y_{t-1}\left[\alpha_y - \alpha_y y_{t-1} - \beta_{xy} x_{t-1} - \beta_{zy} z_{t-1}\right] + \epsilon_{y,t} \\ z_t = z_{t-1}\left[\alpha_z - \alpha_z z_{t-1} - \beta_{xz} x_{t-1} - \beta_{yz} y_{t-1}\right] + \epsilon_{z,t} \end{cases}$$

where $\alpha_x = 3.6$, $\alpha_y = 3.72$, and $\alpha_z = 3.68$. The noise terms $\epsilon_{i,t}(i, =, xyz)$have zero mean and a standard deviation of 0.001. By varying the coupling coefficients ($\beta_{yx}$, $\beta_{zx}$, $\beta_{xy}$, $\beta_{zy}$, $\beta_{xz}$, and $\beta_{yz}$), we generated five network motifs: direct links, causal chains, feedback loops, common drivers (confounding), and colliders. We computed the SCM causal strength from $x$ to $y$ ($\rho_{x\to y}$) across these systems and compared the results with those of CCM and PCM. At a threshold of $T = 0.5$, all methods accurately captured direct links and collider structures. However, PCM failed to distinguish common-driver effects, and CCM struggled to infer causality across diverse topologies (e.g., indirect chains and feedback loops). In contrast, SCM not only effectively eliminated indirect influences but also resolved confounding effects and feedback loops. These results demonstrate that SCM provides a more accurate and robust inference of true direct causal associations by effectively reducing spurious links.

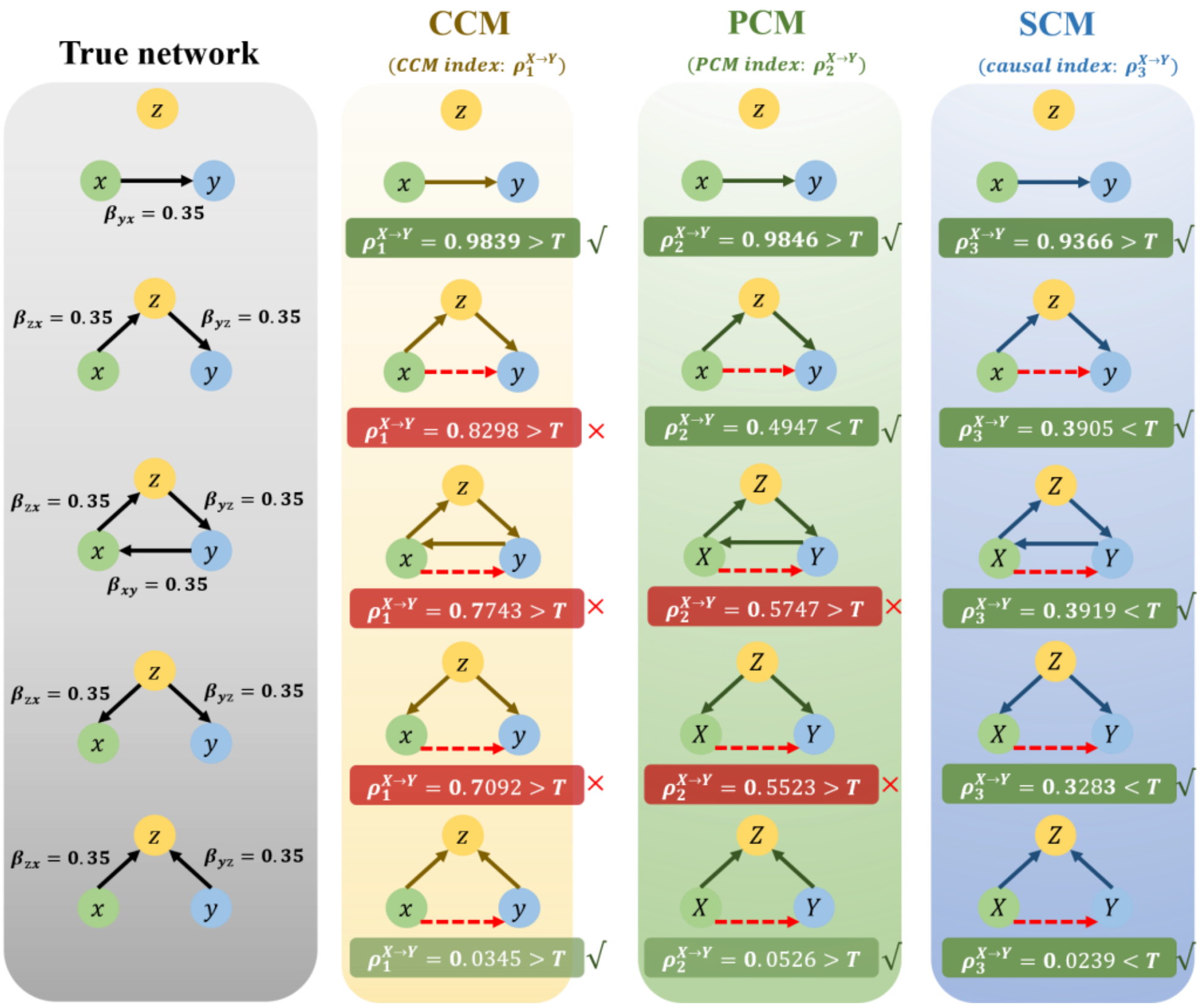

**Fig. 2. Distinguishing direct and indirect causal links across five three-variable motifs.** From top to bottom, the five motifs represent a direct x-to-y link with an independent variable, an indirect mediated chain, an indirect link within a feedback loop, a common driver, and a collider. Columns show the ground-truth networks and causal strengths inferred by CCM, PCM, and SCM, respectively. Red crosses denote incorrectly inferred causal edges, while green checkmarks indicate correctly inferred edges.

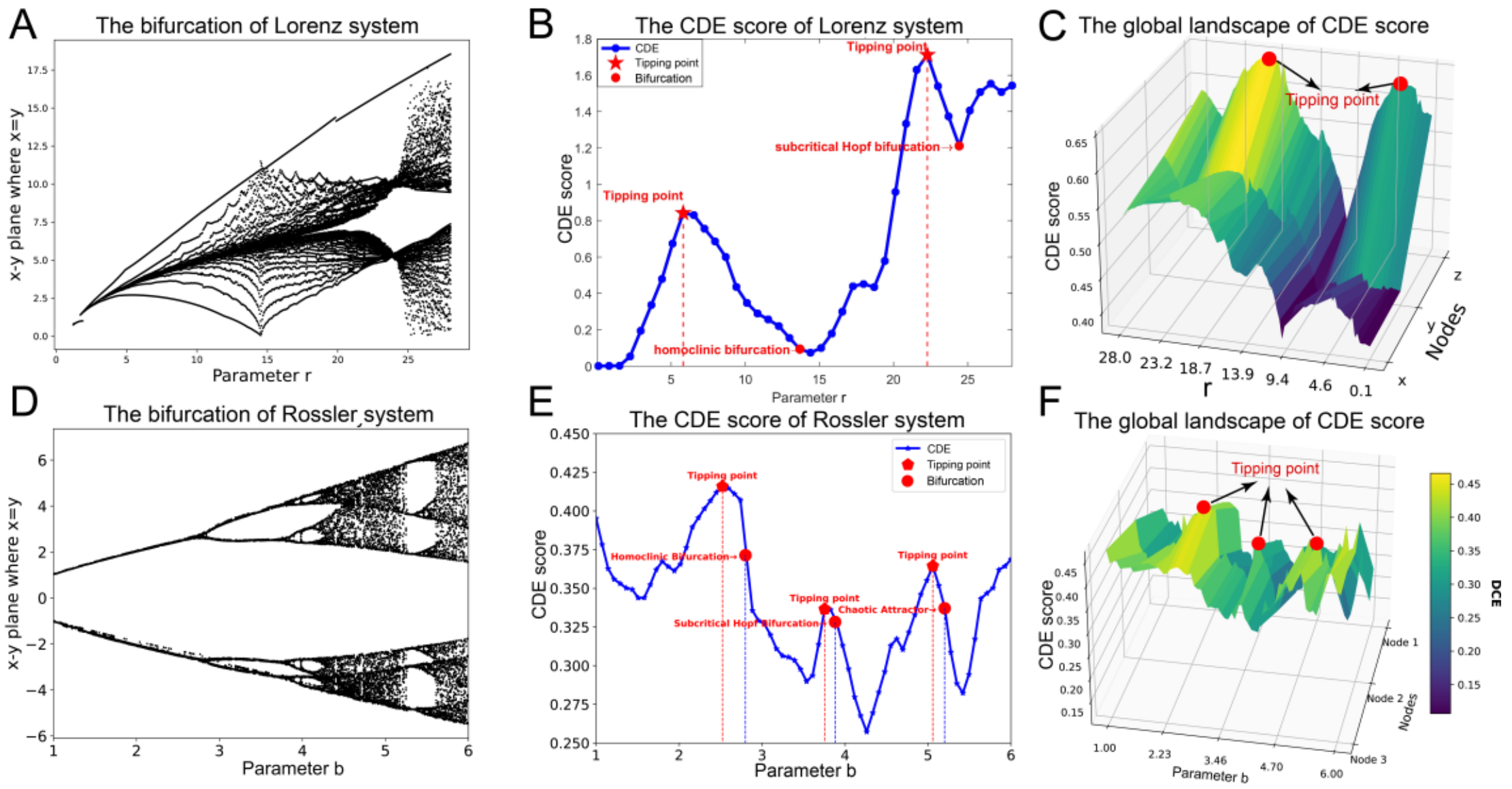

**Fig. 3. Early-warning signals in Lorenz and Rössler systems.** A. Bifurcation diagram of the Lorenz system as the control parameter r varies. B. The curve of CDE score of the Lorenz system. C. The global landscape of CDE score of the Lorenz system. D. Bifurcation diagram of the Rössler system as the control parameter b varies. E. The curve of CDE score of the Rössler system. F. The global landscape of CDE score of the Rössler system.

Beyond causal network reconstruction, we further evaluated the ability of SCM in identifying critical transitions of dynamical systems. Two representative nonlinear systems, Lorenz and Rössler, were examined under varying control parameters The dynamic equations of the Lorenz system are given as follows:

$$\begin{cases} \dot{x} = \sigma(y - x), \\ \dot{y} = rx - y - xz, \\ \dot{z} = xy - bz. \end{cases} \tag{18}$$

where $\sigma = 10$, $b = 8/3$, and the parameter $r$varies within the range of $[0.1, 28]$. As $r$ increases, the system exhibits dynamical transitions near $r = 13.93$ and $r = 24.06$ (Fig. 3A). Notably, the global CDE displays two pronounced peaks at approximately $r = 5.82$ and $r = 22.28$, preceding the corresponding transitions (Fig. 3B). These peaks are characterized by sharp increases followed by subsequent declines, reflecting substantial fluctuations in inferred causal interactions. Furthermore, the node-level CDE landscape reveals heterogeneous responses among system variables, with node $z$exhibiting more pronounced fluctuations than node $y$(Fig. 3C), highlighting the potential of CDE to identify variables contributing strongly to transition-associated network changes. We further examined the Rössler system, described by

$$\begin{cases} \dot{x} = -(y + x), \\ \dot{y} = \dot{x} + ay, \\ \dot{z} = xz - bz + c. \end{cases} \tag{19}$$

where $a = 0.2$and $c = 0.2$, and the parameter $b$ varies within the range of $[1, 6]$. Similarly, the Rössler system exhibited three prominent CDE peaks associated with changes in its dynamical behavior (Fig. 3D and E). CDE successfully identified early-warning signals prior to the first bifurcation ($b \approx 2.8$), the second bifurcation ($b \approx 4.2$), and the chaotic attractor ($b \approx 5.7$) (Fig. 3D). The node-level CDE landscape further revealed distinct fluctuations among system variables (Fig. 3F), indicating that their contributions to the global warning signal are heterogeneous. These findings suggest that changes in inferred causal interactions can provide precursor signals for dynamical transitions along parameter sweeps, while also identifying the variables that contribute disproportionately to the network-level response.

## 3.2. Network reconstruction and critical early warning in an eight-node regulatory network

To further evaluate the performance of SCM in higher-dimensional dynamical systems, we applied it to an eight-node gene regulatory network governed by Michaelis–Menten kinetics (Fig. 4). The system dynamics are described b

$$\begin{cases} \dfrac{dz_1(t)}{dt} = (-3.9|p|)z_1(t) + \zeta_1(t) \\ \dfrac{dz_2(t)}{dt} = -3.5z_2(t) + \zeta_2(t) \\ \dfrac{dz_3(t)}{dt} = \dfrac{(1.267 - 1.365|p|)z_1(t)}{(1 + z_1(t))} + \dfrac{0.042z_2(t)}{(1 + z_2(t))} - 3.62z_3(t) + \zeta_3(t) \\ \dfrac{dz_4(t)}{dt} = \dfrac{0.0875z_2(t)}{(1 + z_2(t))} - 3.75z_4(t) + \zeta_4(t) \\ \dfrac{dz_5(t)}{dt} = \dfrac{0.0037}{(1 + z_1(t))} + \dfrac{0.0037}{(1 + z_2(t))} + \dfrac{0.0105z_3(t)}{(1 + z_3(t))} - 3.65z_5(t) + \zeta_5(t) \\ \dfrac{dz_6(t)}{dt} = \dfrac{0.0123}{(1 + z_1(t))} + \dfrac{0.0123}{(1 + z_2(t))} + \dfrac{0.035z_3(t)}{(1 + z_3(t))} - 3.72z_6(t) + \zeta_6(t) \\ \dfrac{dz_7(t)}{dt} = \dfrac{0.0064}{(1 + z_1(t))} + \dfrac{0.0064}{(1 + z_2(t))} + \dfrac{0.0183z_3(t)}{(1 + z_3(t))} + \dfrac{0.0525}{(1 + z_6(t))} - 3.57z_7(t) + \zeta_7(t) \\ \dfrac{dz_8(t)}{dt} = \dfrac{0.0017}{(1 + z_1(t))} + \dfrac{0.0017}{(1 + z_2(t))} + \dfrac{0.0049z_3(t)}{(1 + z_3(t))} + \dfrac{0.014}{(1 + z_6(t))} - 3.68z_8(t) + \zeta_8(t) \end{cases} \tag{20}$$

where $p$ is a scaling control parameter, $\zeta_i(t)$ $(i = 1, \dots, 8)$ denotes zero-mean Gaussian noise with covariance $k_{ij} = \mathrm{Cov}(\zeta_i, \zeta_j)$, and $z_i(t)$ represents the concentration of the $i$-th moleular species. The degradation rates are $R = (3.9|p|, 3.5, 3.62, 3.75, 3.65, 3.72, 3.57, 3.68)$, and the equilibrium point is $\bar{Z} = (\bar{z}_1, \bar{z}_2, \dots, \bar{z}_8) = (0, 0, \dots, 0)$. By discretizing the differential equations using the Euler method and calculating the eigenvalues, we obtained eight distinct eigenvalues ($0.0302^{|p|}$, 0.0282, 0.0275, 0.0268, 0.0260, 0.0252, 0.0242, 0.0235). The equilibrium point $\bar{Z}$ is stable when $p \in (0, 1]$. As $p \to 0$, the dominant eigenvalue approaches 1, indicating a bifurcation at $p = p_c$ ($p_c \to 0$) where a critical transition occurs.

Fig. 4A shows the ground-truth topology of this eight-node network, while Fig. 4B presents the causal strength matrix inferred by SCM. At a threshold of 0.5, SCM successfully identified many of the reference causal interactions, achieving an area under the receiver operating characteristic curve (AUROC) of 0.9917 across different thresholds. These results demonstrate its effectiveness in reconstructing directed regulatory interactions in this simulated network. Furthermore, Fig. 4C displays the global CDE scores as the system approaches the bifurcation point. As the control parameter $p$approaches $p_c$, the CDE scores exhibit drastic fluctuations revealing that the critical early-warning signal emerges prior to the bifurcation (i.e., at $p = 0$). To further uncover the causal information flow during this critical phenomenon, Fig. 4D illustrates the time-evolving incoming and outgoing CDE for each node. The first subplot provides the average CDE across all nodes as a reference level, while the subsequent eight subplots show the node-specific dynamics. Notably, Nodes 1 and 3 are highly sensitive to the bifurcation, presenting more pronounced critical signals. Specifically, Node 3 shows prominent fluctuations in outgoing entropy at the critical point, whereas Nodes 4, 5, and 8 exhibit more noticeable fluctuations in incoming entropy. These heterogeneous patterns highlight distinct contributions of individual nodes to causal network evolutions and identify candidate dynamical causal markers associated with critical changes.

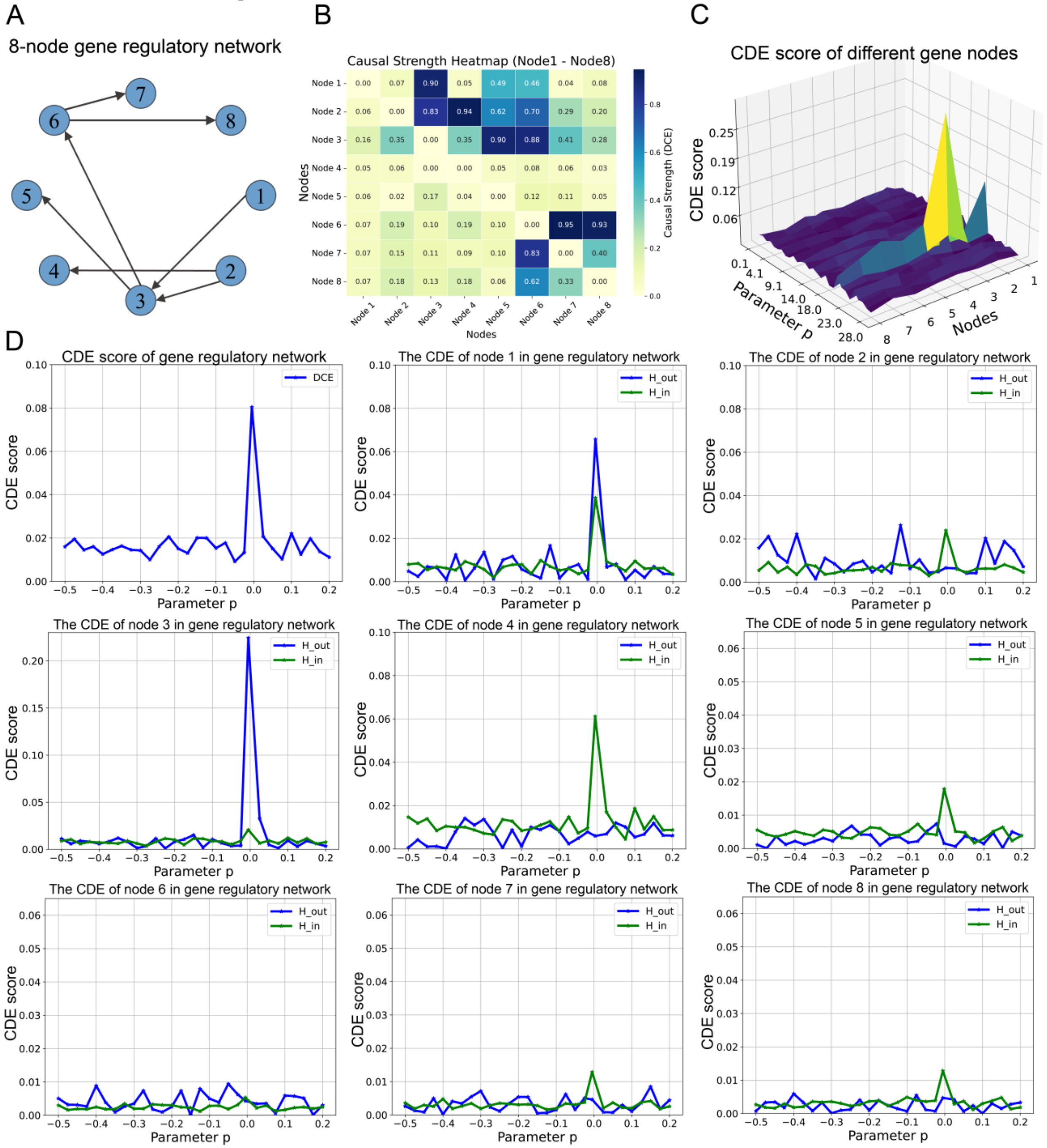

**Fig. 4. Network reconstruction and critical dynamics in an eight-node regulatory network.** A. Ground-truth topology of the eight-node gene regulatory network. B. Causal strength matrix inferred by SCM. C. The global landscape of CDE score across the control parameter p. D. Network-average CDE score (first panel) and node-specific incoming [42] and outgoing (blue) CDE scores. The decomposition reveals heterogeneous node-level contributions near $p = 0$: Nodes 1 and 3 are highly sensitive to the critical transition, with Node 3 exhibiting prominent outgoing fluctuations, whereas Nodes 4 and 5 show more noticeable incoming responses.

## 3.3. Early warning of cell-fate transitions in simulated single-cell differentiation

To further evaluate the effectiveness of SCM to detect critical transitions in complex biological systems, we applied it to a simulated single-cell RNA-sequencing (scRNA-seq) dataset generated using SERGIO [39], a stochastic differential equation-based simulator. The dataset comprised 700 cells, 100 genes, and seven cell types. Principal component analysis (PCA) and trajectory inference were used to construct the dynamic pseudotime trajectory, which revealed a fate transition from Cell type 1 to Cell types 6 and 7 (Fig. 5A). A bifurcation occurs at State 3, followed by two distinct branches that differentiate into Cell types 6 and 7. Fig. 5B illustrates the fluctuations in the causal embedding entropy (CDE) scores during differentiation. A critical point is identified at State 2, anticipating the upcoming state transition. This aligns closely with the observed cell fate transition (Fig. 5A), demonstrating that SCM can provide early warning of cellular fate decisions. Fig. 5C further reveals gene-level pattern changes prior to the bifurcation from a causal perspective. The first subplot displays the incoming CDE for each gene across different differentiation stages, while the second subplot displays the outgoing CDE from each gene. Starting from the critical stage, pronounced fluctuations emerge in the causal entropy of genes within the regulatory network, marking the onset of critical early-warning signals at the molecular level. Together, these results demonstrate the capability of SCM to capture cell fate transitions, validating its accuracy and effectiveness in complex differentiation processes.

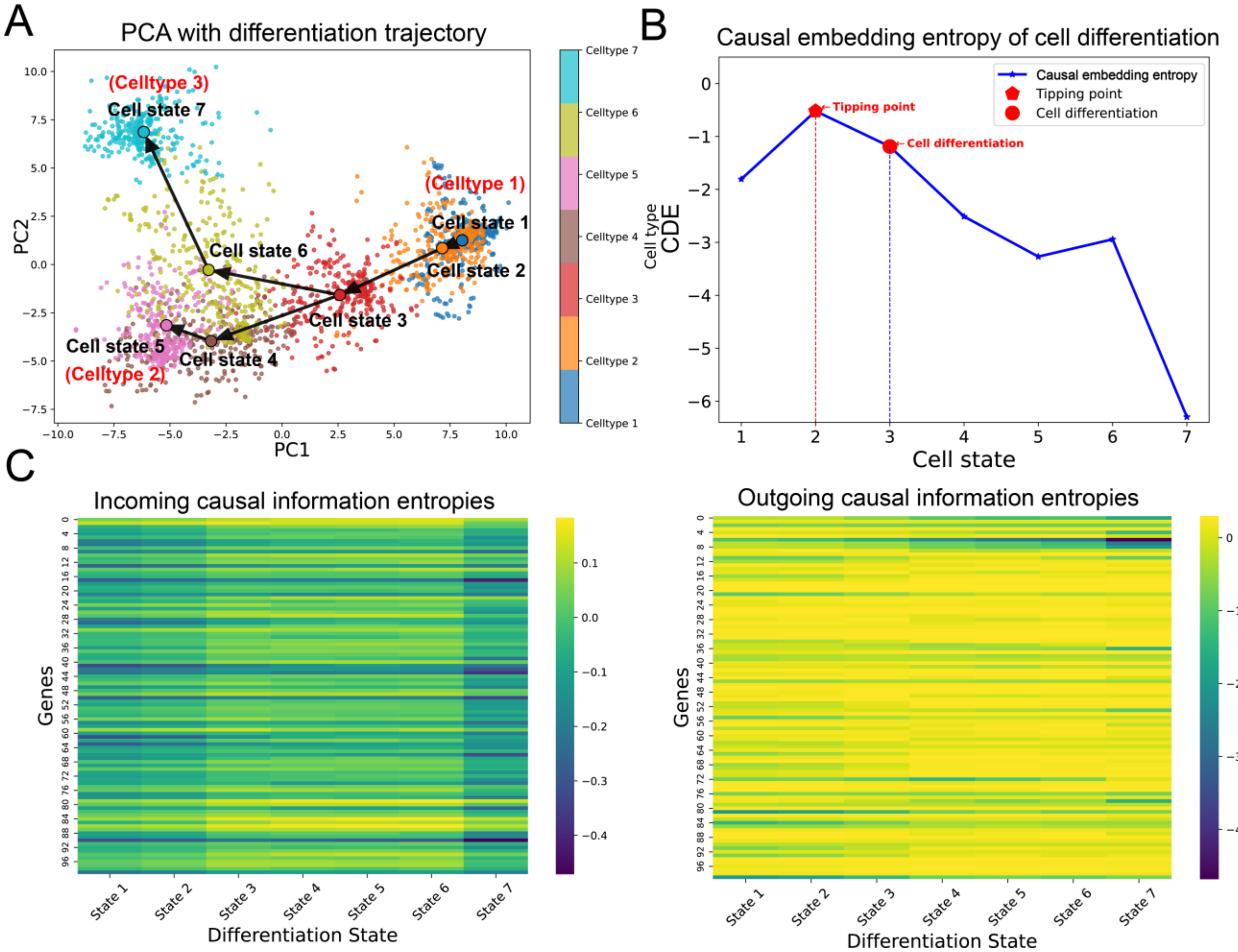

**Fig. 5. Critical transition detection in simulated single-cell differentiation.** A. PCA of the SERGIO dataset (700 cells, 100 genes, seven cell types). Colors distinguish cell-type labels, and the black trajectory indicates the inferred progression through the branching region into two routes ending at Cell types 2 and 3. B. The CDE scores across differentiation states, showing a peak prior to the bifurcation. C. Gene-level heatmaps of incoming (left) and outgoing (right) CDE scores across differentiation states.

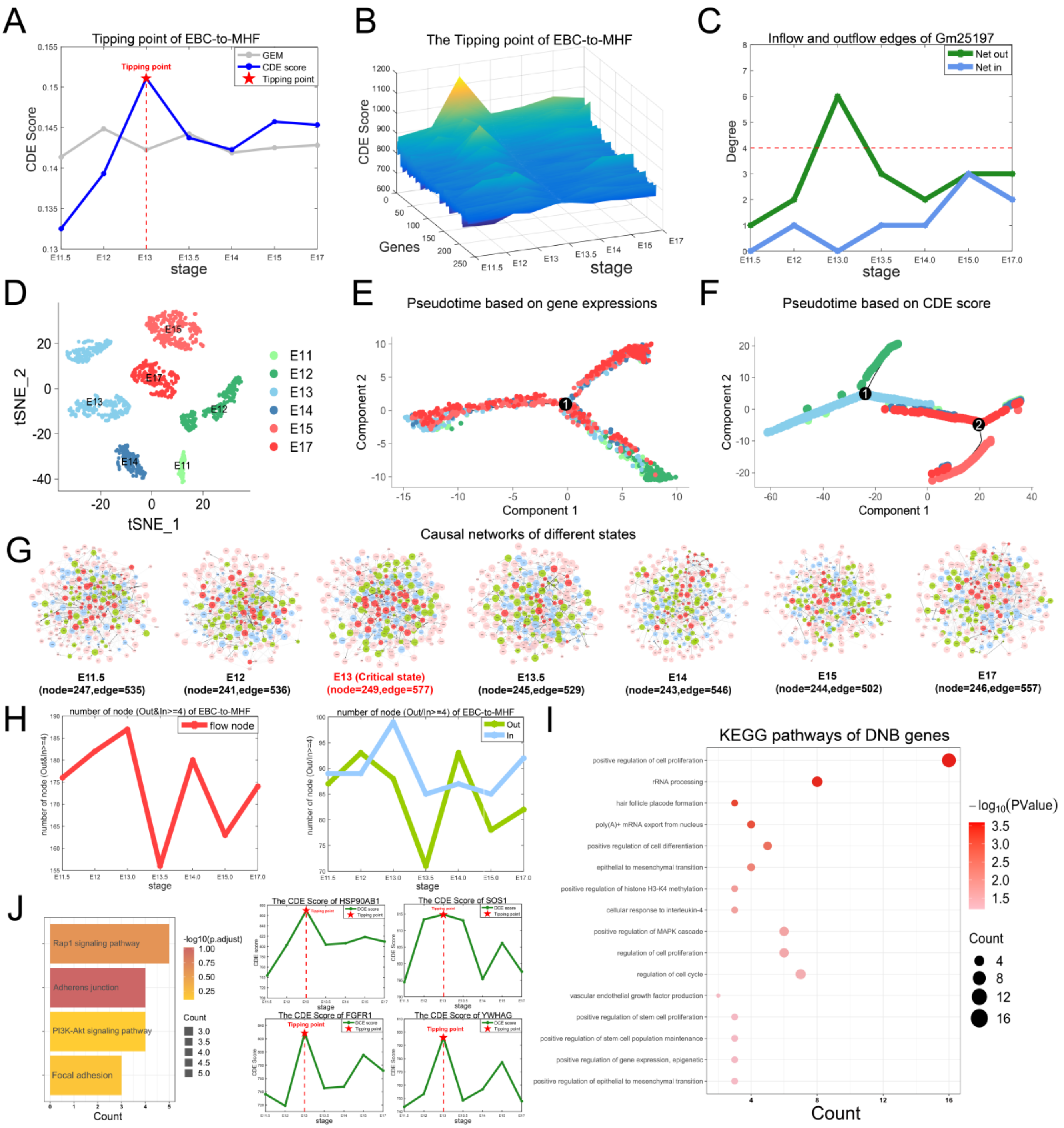

**Fig. 6. Critical states and causal networks during mouse hair-follicle development.** A. Network-level CDE and mean gene expression (GEM) across developmental stages, showing a CDE peak at E13 before the reference stage E13.5. B. The global landscape of CDE score across developmental stages. C. Incoming and outgoing network degrees of Gm25197. D. t-SNE plot of single cells based on CDE features, with colors indicating developmental groups. E, F. Pseudotime projections based on gene expression (E) and CDE (F). G. The dynamic evolution of the causal network during differentiation. Red, green, blue, and pale-pink nodes represent nodes with high bidirectional connectivity, predominantly outgoing connectivity, predominantly incoming connectivity, and non-fluctuating activity, respectively. H. Network degree of the dynamic causal markers across seven stages (first subplot) and the corresponding in- and out-degrees at each stage (second subplot). I. Gene Ontology (GO) biological-process enrichment of candidate marker genes. J. KEGG pathway enrichment (left) and the CDE levels of the four genes (HSP90AB1, SOS1, YWHAG, and FGFR1) involved in the PI3K-Akt signaling pathway.

### 3.4. Unveiling Causal Networks and Critical States During Mouse Hair-Follicle Differentiation

To further evaluate the capability of SCM in inferring causal networks and detecting critical points in real biological datasets, we applied it to a single-cell dataset of epithelial basal cells differentiating toward mouse hair follicles (EBC-to-MHF). The results demonstrated that SCM effectively detected early-warning signals of cell fate determination during embryonic development. As shown by the blue curve in Fig. 6A, the CDE increased sharply at embryonic day 13 (E13), preceding the observation at E13.5 when epithelial basal cells are induced to differentiate into hair follicle stem cells [40]. In contrast, the grey curve in Fig. 6A, representing the temporal fluctuations in gene expression, revealed no visible warning signals before differentiation. This indicates that SCM further uncovers causal network information embedded in gene expression, which is crucial for characterizing early-warning signals in dynamical systems. The CDE landscape further revealed heterogeneous fluctuations among dynamical causal markers (DCMs) near the developmental transition (Fig. 6B). In particular, *Gm25197* exhibited a marked increase in outgoing network degree before follicular induction, without a comparable change in incoming degree (Fig. 6C), indicating a shift in its inferred regulatory connectivity, and the incoming and outgoing causal entropy curves of the other 12 DCMs are shown in Fig. S2. To further assess the utility of CDE in pseudotime analysis, t-SNE visualization based on cell-level CDE revealed distinct clusters across six developmental stages (Fig. 6D). Furthermore, we visualized the CDE scores of nine DCMs across different cell states based on the t-SNE embedding (Fig.S1). These genes exhibited markedly high CDE scores specifically within the E13 cell population, indicating that they underwent specific causal fluctuations at the critical state identified at E13. Compared with gene expression-based pseudotime analysis, CDE-derived trajectories exhibited more clearly defined differentiation paths (Fig. 6 E and F), highlighting its effectiveness in characterizing cell state transitions.

To dissect the dynamic evolution of the causal network during differentiation, we constructed stage-specific network (Fig. 6G) and quantified the information flow of dynamic causal marker genes across different stages. The distribution of highly connected genes and their incoming and outgoing connectivity varied substantially across developmental stages, with pronounced changes around E13 (Fig. 6H). The first subplot in Fig. 6H illustrates the degree of the dynamic causal marker network across seven stages, while the second subplot presents the in- and out-degrees at each stage. Overall, the in-degree (i.e., incoming CDE) of the network dictates the emergence of critical signals to a greater extent. To evaluate the functional roles of identified dynamical causal markers (DCMs) in cell differentiation, we performed GO and KEGG enrichment analyses. Fig. 6I shows that the marker genes were enriched in biological processes closely related to cell proliferation and differentiation, hair-follicle placode formation, epithelial-mesenchymal transition, as well as epigenetic and signaling events (e.g., histone H3-K4 methylation, MAPK cascades, stem cell maintenance). Certain marker genes were also enriched in critical signaling pathways during cell differentiation, such as PRKD3, GNAS, ACTB, FGFR1, and ACTG1 in the Rap1 signaling pathway, and HSP90AB1, SOS1, YWHAG, and FGFR1 in the PI3K-Akt signaling pathway (Fig. 6J). The four subplots in the right panel of Fig. 6J further illustrate the CDE levels of the four genes involved in the PI3K-Akt signaling pathway. Their CDE increased significantly before epithelial basal cells were induced to differentiate into hair follicle stem cells, and then stabilized or declined after differentiation, marking the emergence of molecular-level critical signals and demonstrating that SCM can effectively detect early-warning signals of cell fate decisions during embryonic development.

### 3.5. Critical transition detection and causal network inference in an Alzheimer's disease progression

Alzheimer's disease (AD) is a progressive neurodegenerative disorder characterized by complex molecular alterations, yet the causal regulatory interactions and critical transitions underlying disease progression remain to be investigated. Here, we applied SCM and causal embedding entropy (CDE) to the AD11 mouse transcriptomic dataset [41] to reconstruct gene regulatory networks and identify potential early-warning signals of neurodegeneration. The dataset contains gene expression profiles from the hippocampus, cortex, and basal forebrain of AD11 mice and transgenic controls at four ages (1, 3, 6, and 15 months). By analyzing changes in inferred causal interactions and CDE, we aimed to identify candidate dynamical causal markers associated with AD-like disease progression.

The global CDE landscape in Fig. 7A exhibits the dynamic evolution of CDE of each gene, in which the CDE values of some genes exhibit an increasing trend at stage 2. Then we further show the CDE of the dynamical causal markers during cell differentiation, there is a sharp increase at stage 2, followed by a decline at subsequent stages, suggesting a potential critical transition preceding cell fate determination at stage 3. Notably, the outflow CDE peaks at stage 2, whereas the inflow CDE remains relatively stable (Fig. 7C), indicating that dynamical causal markers primarily act as upstream regulators during the critical transition of neurodegeneration. Fig. 7D further illustrates dynamic fluctuations of the causal networks among dynamical causal markers. At stage 2, the network exhibits markedly stronger causal interactions and denser connectivity, particularly among Sorl1, Gria2, Celf1, Ptk2b, Keap1, Ldlr, Gria1, Syn2, and Grin2b. Notably, Sorl1 emerges as a prominent upstream node with directed causal links to Gria2, Ptk2b, and Celf1.

Sorl1 is associated with neuronal endosomal trafficking and synaptic receptor recycling [43, 44], while other identified genes are involved in glutamatergic signaling, synaptic function, and plasticity [45-47]. In addition, Ptk2b has been implicated in synaptic dysfunction associated with Alzheimer's disease [48]. Fig. 7E further display the inflow and outflow CDEs at different stages, there is a marked changes in outflow CDE of Ptk2b and Grin2b, highlighting their potential roles as dynamical causal markers. Fig. 7E further displays the inflow and outflow CDEs at different stages, in which Ptk2b and Grin2b exhibit pronounced increases in outflow CDE at stage 2, indicating their upstream regulatory roles at the critical stage. These findings highlight the reorganization of neuronal regulatory interactions at the candidate critical stage. In addition, the identified dynamical causal markers were also enriched in several biological processes, e.g., Grin2b, Ptk2b, Gria1, and Gria2 were involved in postsynaptic specialization, Grin2b and Ptk2b in synaptic plasticity, and Gria1 and Gria2 in receptor-mediated endocytosis. And the KEGG analysis (Fig. 7G) further revealed that Gria1, Gria2, Grin2a, and Grin2b were enriched in glutamatergic synapse and long-term potentiation pathways, whereas Grin2b, Mapt, and Lrp1 were associated with the AD pathway. These findings highlight the crucial roles of the identified dynamical causal markers in synaptic dysfunction and neurodegeneration-related processes in Alzheimer's disease.

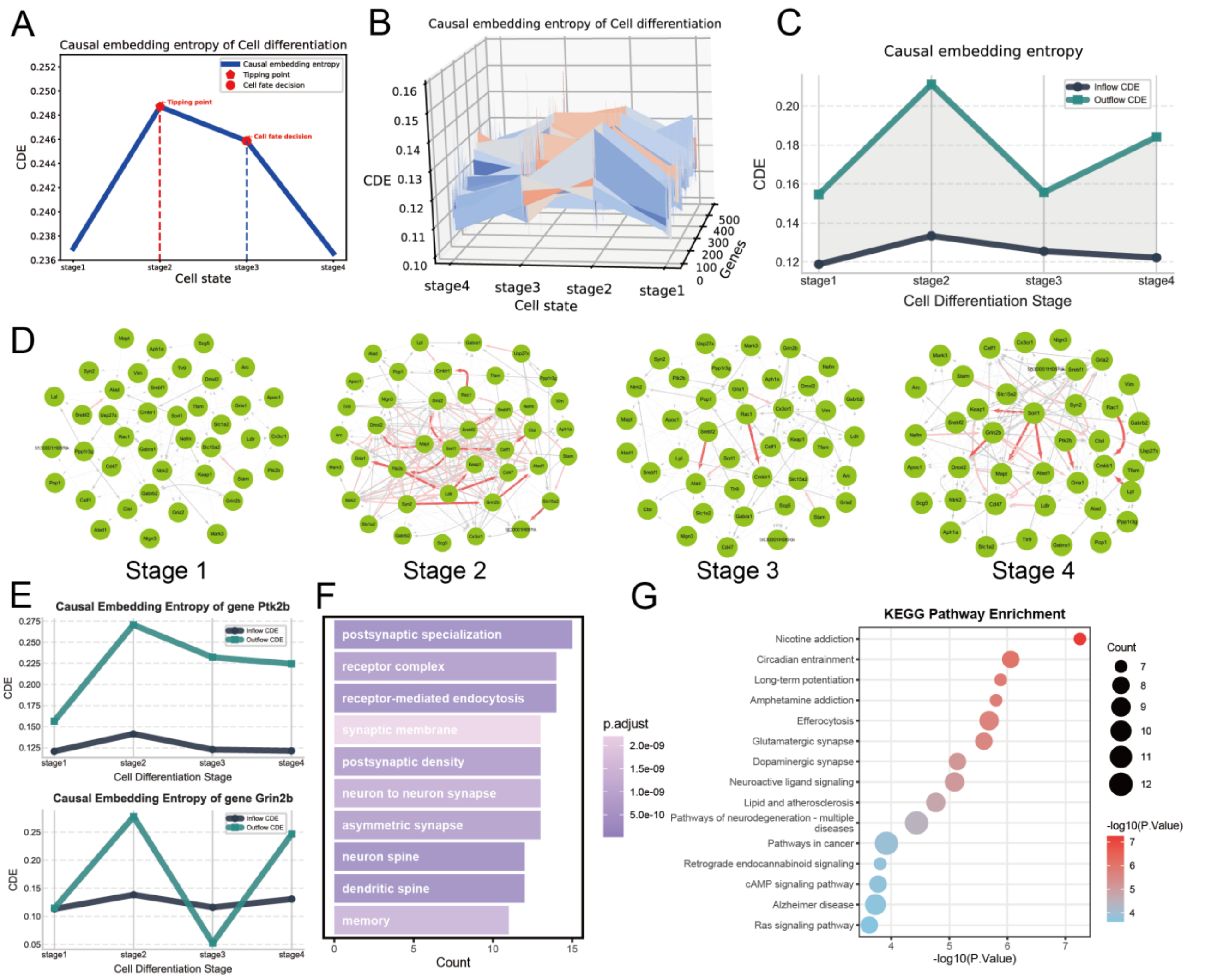

**Fig. 7. Characterizing critical states and causal network remodeling during mouse hair-follicle development.** A. Network-level CDE and mean gene expression across developmental stages, showing a CDE peak at E13 before the reference stage E13.5. B. The global landscape of CDE score across developmental stages. C. Network-level incoming and outgoing CDE profiles. D. Stage-specific inferred causal subnetworks from Stage 1 to Stage 4. E. Incoming and outgoing CDE profiles of *Ptk2b* and *Grin2b* (bottom). F. Gene Ontology (GO) enrichment of candidate marker genes. G. KEGG pathway enrichment of candidate marker genes.

# 4. Discussion

In various fields, accurately predicting critical transitions or bifurcations is essential for understanding system dynamics. In complex biological systems—such as cellular differentiation, embryonic development, and disease progression—critical transitions are often accompanied by subtle changes in molecular interactions. Therefore, reconstructing real regulatory networks and capturing their dynamic reorganization are crucial for comprehending underlying biological mechanisms. However, two major challenges remain. However, two major challenges remain: (i) nonlinear couplings often cause indirect effects to be misdiagnosed as direct causality; and (ii) conventional early-warning methods rely on undirected correlations, lacking the directional logic to pinpoint upstream drivers during critical states.

To address these challenges, we developed SCM, a computational framework that integrates delay-coordinate embedding, S-map-based locally weighted regression, and partial correlation analysis to reconstruct direct dynamical causal networks from observed time series. Building on the inferred networks, we further proposed Causal Embedding Entropy (CDE) to quantify directed causal information flow and identify Dynamical Causal Markers (DCMs), thereby establishing a unified framework that bridges dynamical causal inference with critical-transition analysis. Applications to simulated dynamical systems, gene regulatory networks, single-cell differentiation, embryonic development, and neurodegeneration demonstrate the potential of this framework to reconstruct causal interactions, identify transition-associated network changes, and reveal candidate dynamical causal markers (DCMs).

The advantages of our framework can be summarized as follows. First, a key methodological advance of SCM lies in its continuous transition from global linearity to local nonlinearity via the localization parameter $\theta$, coupled with a two-stage sequential cross-mapping procedure that explicitly filters out indirect dependencies, enabling direct causal network inference and more accurate biological network reconstruction. Sceond, based on the directionality of the causal network, this method can further quantify information flow through incoming and outgoing entropy, thereby characterizing the regulatory roles of individual nodes within the network. Third and importantly, this work further investigates how causal strengths evolve as the dominant eigenmode approaches instability. The theoretical analysis shows that causal interactions associated with dominant and non-dominant variables exhibit distinct changes near critical transitions. These findings provide a theoretical motivation for extending critical-state analysis from molecular correlations to causal regulation networks. Thus, the central contribution of our framework is to bridges dynamical causal inference and critical-state detection grounded in critical slowing down through the dynamic evolution of causal networks.

It is worth noting that there remain several limitations that merit further investigation. First, the theoretical framework relies on accurate state-space reconstruction based on Takens' embedding theorem, which fundamentally requires continuous time-series data. This prerequisite limits its direct applicability to non-sequential datasets. Second, the current SCM framework eliminates indirect associations by conditioning on all observed variables. Consequently, it cannot inherently resolve confounding effects induced by unobserved variables. Third, because the link between causal strength fluctuations and critical transitions depends on statistical properties of reconstructed variables and behavior of dominant dynamical modes, the CDE response is inherently system-dependent. Future work should thus explore causal inference under partial observability and establish more general theoretical conditions.

# Data availability

All datasets utilized by this work are obtained from publicly accessible databases. The single-cell RNA-sequencing dataset of mouse hair-follicle development is available under accession number GSE147372 [40], whereas the microarray dataset from the AD11 anti-NGF transgenic mouse model of neurodegeneration is available under accession number GSE63617 [41].